\documentclass[
]{ceurart}

\usepackage{listings}
\begin{document}

\copyrightyear{2024}
\copyrightclause{Copyright for this paper by its authors.
  Use permitted under Creative Commons License Attribution 4.0
  International (CC BY 4.0).}

\conference{GeoExT 2024: Second International Workshop on Geographic Information Extraction from Texts at ECIR 2024, March 24, 2024, Glasgow, Scotland}

\title{Do Sentence Transformers Learn Quasi-Geospatial Concepts from General Text?}

\author[1]{Ilya Ilyankou}[
    email=ilya.ilyankou.23@ucl.ac.uk,
    orcid=0009-0008-7082-7122
]
\cormark[1]
\address[1]{SpaceTimeLab, University College London, UK}

\author[1]{Aldo Lipani}[
    email=aldo.lipani@ucl.ac.uk,
    orcid=0000-0002-3643-6493
]

\author[2]{Stefano Cavazzi}[
    email=stefano.cavazzi@os.uk,
    orcid=0000-0003-3575-0365
]
\address[2]{Ordnance Survey, Southampton, UK}

\author[1]{Xiaowei Gao}[
    email=xiaowei.gao.20@ucl.ac.uk,
    orcid=0000-0003-3273-7499
]

\author[1]{James Haworth}[
    email=j.haworth@ucl.ac.uk,
    orcid=0000-0001-9506-4266
]

\cortext[1]{Corresponding author.}

\begin{abstract}
    Sentence transformers \cite{reimers_sentence-bert_2019} are language models designed to perform semantic search.
    This study investigates the capacity of sentence transformers, fine-tuned on general question-answering datasets
    for asymmetric semantic search, to associate descriptions of human-generated routes across Great Britain
    with queries often used to describe hiking experiences. We find that sentence transformers have some zero-shot
    capabilities to understand quasi-geospatial concepts, such as route types and difficulty, suggesting their
    potential utility for routing recommendation systems.
\end{abstract}


\maketitle

\section{Introduction}

Semantic search is different from traditional, keyword search in that it tries to capture
the \textit{intent} of the searcher \cite{wei_search_2008}.
For a good semantic search system, the sentence \textit{`The fox sat on the mat'}
should be similar to \textit{`An animal rested upon the rug'}, but different from
\textit{`Fox News on Sat'}.

\href{https://www.sbert.net/}{SentenceTransformers} \cite{reimers_sentence-bert_2019} is a Python library that contains a collection
of predominantly BERT-family \cite{devlin_bert_2019} transformer-based neural network models that are fine-tuned for semantic search.
This study investigates the extent to which asymmetric semantic search models, designed for shorter queries and longer
documents \cite{zhang_question_2014}, fine-tuned on general (non-geospatial) question-answering datasets, can understand vague,
subjective, and complex quasi-geospatial concepts. For example, do such models consistently associate a query \textit{`a walk with a variety of landscapes'} with documents describing longer walks going through various terrains over the documents
describing shorter, solely urban or rural walks?

This topic is important because when searching for (hiking) activities, people tend to describe their physical abilities or desired experiences \cite{davies_who_2018, molokac_preference_2022, hull_landscape_1995} over using precise geospatial terms.

\section{Methodology}

In our experiment, we take existing user-generated routes across Great Britain,
add geospatial context to generate their textual descriptions,
use Sentence Transformers to create vector embeddings for each description,
and compare these vectors with embeddings of user queries.

\subsection{Data}
We use user-generated hiking routes from the Ordnance Survey's OS Maps app—specifically, a subset of 501,294 routes classified by Ballatore et al. \cite{ballatore_context_2023}.
We remove very short (under 1 km) and very long (over 50 km) routes to focus on those
that would typically be of interest to leisure hikers, and can be completed in a day.
After further removing routes with obvious GPS signal issues, we have 496,723 routes whose average length is 11,289 metres.
We then assign a set of attributes to each route using \texttt{geopandas} \cite{noauthor_geopandas_2024}
and several Ordnance Survey datasets. A full list of attributes is shown in Table \ref{tab:route_attributes} in Appendix.

\subsection{Generating descriptions}
We use simple template language to generate 3-4 sentence descriptions for each route based on route attributes.
Descriptions mention length, shape, start and end points, total elevation gain and steepness.
We explicitly state the walk is \textit{`predominantly uphill'} or \textit{`predominantly downhill'} where the total elevation gain exceeds (or is less than) the total elevation loss by at least 100 metres; this categorisation applies to approximately 8\% of all routes.
The descriptions also include the types of areas the routes go near or through, including coast, 
surface water, woodland, green space, urban areas, and national parks. For each area type, we mention for how
long in percentages, swapping numbers (e.g., `60 percent') for words (e.g., `sixty percent' or `most')
roughly half the time. We do this because language models are known to under-perform when
required to work with numbers \cite{petrak_arithmetic-based_2023}. 

The resulting descriptions are between 112 and 589 characters long, with mean and median of 299 and 296 characters respectively.
The longest description reads:

\begin{quote}
    `This is a twenty-five km walk that begins in Yelverton, West Devon, Devon and ends in Plymouth, City of Plymouth. Total elevation gain is seven hundred and thirty-nine metres, and elevation grade is 2.9. The walk is predominantly downhill. About 25 percent of the walk is within a national park, about thirty percent of the walk is in a wooded area, about thirty-three percent of the walk goes through an urban area, about 12 percent of the walk is within green space, about twenty percent of the walk is along the coast, about forty-three percent of the walk is alongside a body of water.'
\end{quote}

Other examples of descriptions of various lengths are shown in Table \ref{tab:descriptions} in Appendix.

\subsection{Matching queries with descriptions}

We calculate vector representations, or embeddings, of all textual descriptions and user queries using \texttt{msmarco-\{MiniLM-L6|distilbert\}-cos-v5} and \texttt{multi-qa-\{MiniLM-L6|distilbert|mpnet-base\}-cos-v1} models. These are based on MiniLM \cite{wang_minilm_2020}, DistilBERT \cite{sanh_distilbert_2020}, and MPNet \cite{song_mpnet_2020} architectures, and fine-tuned on MS MARCO \cite{bajaj_ms_2018} (about 500k records) and/or a compilation of question-answering datasets which we refer to here as Multi-QA \cite{noauthor_sentence-transformersmulti-qa-minilm-l6-cos-v1_2024} (about 215M records). Neither collection is specific to the geospatial domain. Models tuned on MS MARCO support input sequences of up to 384 tokens (word pieces), and those tuned on Multi-QA support up to 512 tokens; our inputs fit comfortably within both limits, with the longest description of 589 characters represented by 129 tokens.

We use 20 queries (see Table \ref{tab:queries} in Appendix) that resemble questions (e.g., \textit{`what is a walk for an expert hiker'}), and calculate cosine similarity between all queries and route descriptions to rank the relevance of each description for each query. The queries are inspired by various research papers that studied hiking experiences \cite{davies_who_2018, hull_landscape_1995, molokac_preference_2022, sarjakoski_analysis_2012, ballatore_context_2023, calbimonte_semantic_2020}.

\subsection{Visualising results}

Our experiment is difficult to assess using standard information retrieval quality metrics such as mean reciprocal rank (MRR) or mean average precision (MAP) \cite{bellogin_statistical_2017} given we cannot easily label an individual route description (document) as relevant to the query or not. Instead, we are interested in the overall patterns of how the documents are ranked.

We decided to plot \textbf{cumulative means} of relevant route attributes for ranked documents,
sorted from best to worst match (one can think of it as plotting \texttt{average@k} of relevant route attribute means for all \texttt{k} between 1 and 496,723). As such, the y-value of the left-most point of each
line chart represents the attribute value of the top-matching document, while the y-value of the
right-most point represents the mean attribute value for the whole dataset.
An increasing cumulative mean (i.e., a line going up) signifies higher-ranking documents
(on the left of the x-axis) typically having lower values than lower-ranking documents (to the right), and vice versa. We utilise a logarithmic scale for the x-axis to highlight the cumulative means of top-ranking documents, while also presenting the overall trend for all documents.

\section{Results}

The results are mixed. Unsurprisingly, sentence transformers do better when
user queries have similar terminology to the documents.
All five models are able to associate a `seaside walk' with routes described as
having longer stretches along the coast. Four models clearly associate `a walk for someone who enjoys
town walks' with routes going through urban areas.

\begin{figure}
  \centering
  \includegraphics[width=14.5cm]{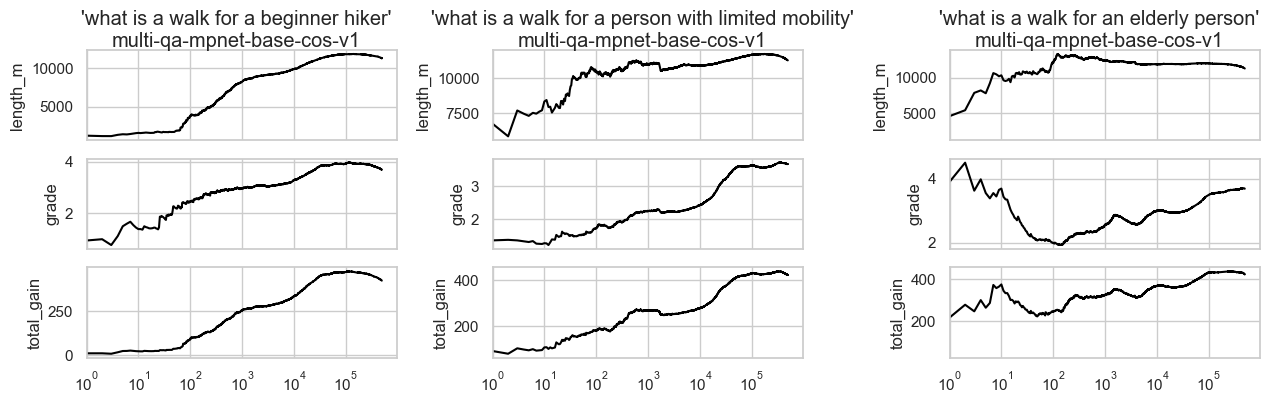}
  \caption{Cumulative mean length, grade, and elevation gain for what should be \textit{easier} routes}
  \label{fig:easy}
\end{figure}

Figure \ref{fig:easy} shows cumulative mean length, grade, and elevation gain for
more complex queries targeting easier walks produced by \texttt{multi-qa-mpnet-base-cos-v1}.
The queries mentioning a `beginner hiker' and a `person with limited mobility' are
indeed associated with shorter and flatter routes; the results are less clear
for the query mentioning an `elderly person'.

\begin{figure}[htp]
  \centering
  \includegraphics[width=14.5cm]{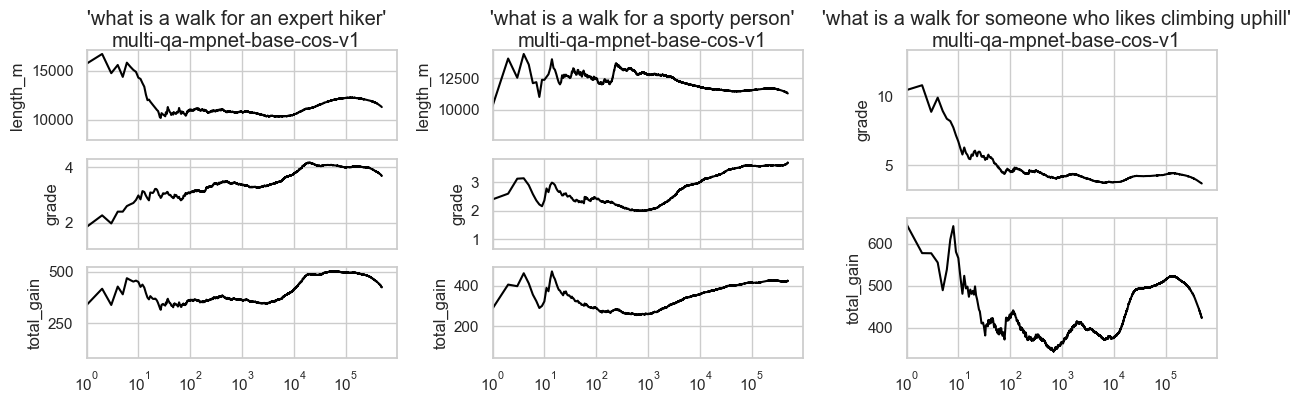}
  \caption{Cumulative mean length, grade, and elevation gain for what should be \textit{harder} routes}
  \label{fig:hard}
\end{figure}

Conversely, Figure \ref{fig:hard} shows results produced by the same model
for queries aimed at more challenging hiking experiences.
While the top-10 or so results for an `expert hiker' are indeed longer walks,
both the slope and total elevation gain patterns are not convincing. A `sporty person' will receive
similarly disappointing suggestions. But `someone who likes climbing uphill' will be pleasantly
surprised, given both the grade and total elevation gain are much higher for best matches.

\begin{figure}[htp]
  \centering
  \includegraphics[width=14.5cm]{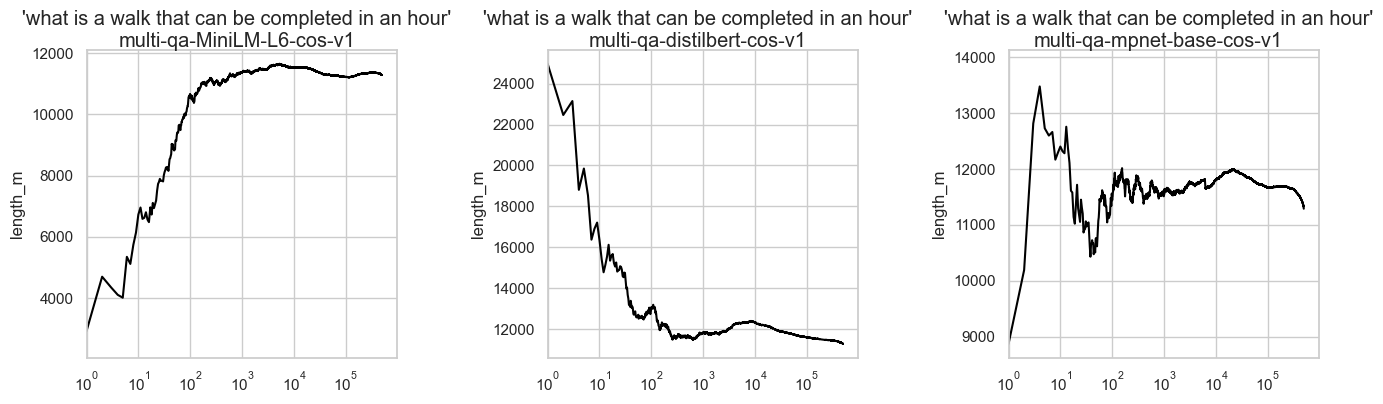}
  \caption{Three models, even though fine-tuned on the same dataset, disagree on which walks can be completed in under 1 hour}
  \label{fig:hour}
\end{figure}

Peculiarly, even when fine-tuned on the same Multi-QA dataset,
MiniLM, DistilBERT, and MPNet models are in total disagreement over the walks that can be completed
in under an hour (Figure \ref{fig:hour}). We generally expect these to be routes of under 5 km \cite{club_scottish_1893}.
While MiniLM shows a logical pattern of top-matching routes being shorter, DistilBERT ranks the results in a near reverse order;
cumulative mean length of routes ranked by MPNet seems to hover around the dataset mean.

None of the models are good at associating `long' and `very long' walks with higher kilometre values.
Regrettably, most models associate `someone seeking greater challenges' with descriptions of shorter and flatter walks,
and queries for people `preferring wilderness to man-made' barely relate to walks going through national parks.
Full results are available in the Appendix.

\section{Conclusion}

Sentence transformers, fine-tuned on a corpus of general question-answer pairs for asymmetric semantic search,
demonstrate some zero-shot ability to associate short and subjective queries looking for particular hiking experiences
with synthetically composed route descriptions. One tested model was able to relate 
`beginner hikers' and those with `limited mobility' to shorter and flatter walks; another was able to
associate `walks that can be completed in under an hour' with shorter walks. Models fine-tuned on the same dataset
sometimes showed very different results, signalling that architectures and pre-training matter.

In future work, a more systematic approach to evaluate sentence transformers and other language models
for geospatial understanding should be introduced. We suggest focusing on four aspects: model architecture,
datasets for fine-tuning, geospatial descriptions, and evaluation. Firstly, a wider array of sentence transformers
should be tested to identify which architectures, BERT-based and beyond, achieve better results.
Secondly, existing general question-answering datasets used for fine-tuning should be evaluated both
independently and in combination to see how dataset size, theme, and quality affect learning. Although the user queries we tested
are not `traditionally' geospatial, will using \textit{smaller} datasets of \textit{primarily geographic} questions
make sentence transformers better understand hiking and other active living experiences? Thirdly, we recognise that using a generic
template to describe routes is just one of many ways to represent geospatial data as text. As such,
we suggest exploring more sophisticated approaches of generating descriptions for routes and other geospatial objects (e.g., specific locations
represented as points, and general areas represented as polygons), given that various sources
(people, websites) describe such objects differently. And lastly, a more formal way of evaluating ranked
results should be explored, accounting for the fact that user queries tested here are more subjective and
incomplete than those used in typical information retrieval tasks.

\clearpage

\begin{acknowledgments}
    This work was supported by the Ordnance Survey,
    and the Engineering and Physical Sciences Research Council [grant no. EP/Y528651/1].
\end{acknowledgments}

\bibliography{articles}

\clearpage

\appendix

\section{Tables}

\begin{table}[htp]
  \caption{Route attributes}
  \label{tab:route_attributes}
  \begin{tabular}{l|p{100mm}|l}
    \toprule
    Attribute&Description&Dataset\\
    \midrule
    \texttt{length\_m}&Route length in metres&-\\
    \texttt{total\_gain}&Total elevation gain (ascent)&OS Terrain 5\\
    \texttt{total\_loss}&Total elevation loss (descent)&OS Terrain 5\\
    \texttt{grade}&Hiking grade, calculated as ($\texttt{total\_gain} \div \texttt{length\_m} \times \texttt{100})$&-\\
    \texttt{is\_circular}&\textit{True} if route start and end points are within 500m of each other&-\\
    \texttt{is\_out\_and\_back}&\textit{True} if route is circular, and its 25m buffer has a high overlap with itself&-\\
    \texttt{start\_place}&Name of the nearest place to the route's start point, limited to 1km&OS Open Names\\
    \texttt{end\_place}&Name of the nearest place to the route's end point, limited to 1 km&OS Open Names\\
    \texttt{along\_surfacewater}&Percent of route length that falls within 50m of a body of water&OS Zoomstack\\
    \texttt{along\_coast}&Percent of route length that lies within 150m of GB boundary&OS Zoomstack\\
    \texttt{is\_coastal}&\textit{True} if at least 50\% of the walk length is along the coast&-\\
    \texttt{in\_national\_parks}&Percent of route length that falls within a national park&OS Zoomstack\\
    \texttt{in\_greenspace}&Percent of route length that falls within a green space boundary&OS Zoomstack\\
    \texttt{in\_woodland}&Percent of route length that falls within a woodland boundary&OS Zoomstack\\
    \texttt{in\_urban}&Percent of route length that falls within an urban area boundary&OS Zoomstack\\
    \bottomrule
\end{tabular}
\end{table}

\begin{table}[htp]
  \caption{Examples of generated descriptions of various character lengths}
  \label{tab:descriptions}
  \begin{tabular}{p{15cm}|r}
    \toprule
    Description&Length\\
    \midrule
    This is a circular, ten km walk that begins and ends in Priddy, Somerset. Total elevation gain is 130 metres, and elevation grade is 1.3.&138\\
    \midrule
    This is a circular, 12 km walk that begins and ends in Tarrant Gunville, Dorset. Total elevation gain is one hundred and ninety-one metres, and elevation grade is 1.6.&168\\
    \midrule
    This is a 2 km coastal walk that begins in Glenuig Bay, Highland and ends in Smirisary, Highland. Total elevation gain is 142 metres, and elevation grade is 4.9. About seven percent of the walk is in a wooded area, about 62 percent of the walk is along the coast.&263\\
    \midrule
    This is a circular, twenty-five km walk that begins and ends in Milton, Stirling. Total elevation gain is 1846 metres, and elevation grade is 7.4. The walk is entirely within a national park, about twenty percent of the walk is in a wooded area, about ten percent of the walk is alongside a body of water.&305\\
    \midrule
    This is a 6 km walk that begins in Pebbly Hill, Cotswold, Gloucestershire and ends in Stow-on-the-Wold, Cotswold, Gloucestershire. Total elevation gain is 215 metres, and elevation grade is 3.5. The walk is predominantly uphill. About seven percent of the walk is in a wooded area, about 8 percent of the walk goes through an urban area.&337\\
    \midrule 
    This is a 22 km walk that begins in Rampart Head, Cumberland and ends in Little Caldew, Cumberland. Total elevation gain is 222 metres, and elevation grade is 1.0. About 6 percent of the walk is in a wooded area, about 9 percent of the walk goes through an urban area, about 7 percent of the walk is within green space, about 18 percent of the walk is along the coast, about twenty-seven percent of the walk is alongside a body of water.&437\\
    \midrule 
    This is a nineteen km walk that begins in Millbrook, Caerffili - Caerphilly and ends in Ynysfro Reservoirs, Casnewydd - Newport. Total elevation gain is seven hundred and twenty-four metres, and elevation grade is 3.7. The walk is predominantly downhill. About seventeen percent of the walk is in a wooded area, about forty-five percent of the walk goes through an urban area, about eight percent of the walk is within green space, about 17 percent of the walk is alongside a body of water.&490\\
    \midrule
    This is a 7 km coastal walk that begins in Rhoose Cardiff International Airport, Rhoose / Y Rhws, Bro Morgannwg - the Vale of Glamorgan and ends in Storehouse Point, Bro Morgannwg - the Vale of Glamorgan. Total elevation gain is 194 metres, and elevation grade is 2.7. About thirteen percent of the walk is in a wooded area, about 19 percent of the walk goes through an urban area, about 17 percent of the walk is within green space, 77 percent of the walk is along the coast, about twelve percent of the walk is alongside a body of water.&539\\
    \bottomrule
\end{tabular}
\end{table}

\begin{table}[htp]
  \caption{User queries}
  \label{tab:queries}
  \begin{tabular}{p{90mm}|p{70mm}}
    \toprule
    Query&Relevant attributes\\
    \midrule
    what is a short walk&\texttt{length\_m}\\
    \midrule
    what is a very short walk&\texttt{length\_m}\\
    \midrule
    what is a long walk&\texttt{length\_m}\\
    \midrule
    what is a very long walk&\texttt{length\_m}\\
    \midrule
    what is a walk by the seaside&\texttt{is\_coastal}\\
    \midrule
    what is a walk through the woods&\texttt{in\_woodland}\\
    \midrule
    what is an urban walk&\texttt{in\_urban}\\
    \midrule
    what is a country walk&\texttt{in\_urban}, \texttt{in\_greenspace}\\
    \midrule
    what is a walk for a beginner hiker&\texttt{length\_m}, \texttt{grade}, \texttt{total\_gain}\\
    \midrule
    what is a walk for an expert hiker&\texttt{length\_m}, \texttt{grade}, \texttt{total\_gain}\\
    \midrule
    what is a walk for a sporty person&\texttt{length\_m}, \texttt{grade}, \texttt{total\_gain}\\
    \midrule
    what is a walk for a person with limited mobility&\texttt{length\_m}, \texttt{grade}, \texttt{total\_gain}\\
    \midrule
    what is a walk for an elderly person&\texttt{length\_m}, \texttt{grade}, \texttt{total\_gain}\\
    \midrule
    what is a walk that can be completed in an hour&\texttt{length\_m}\\
    \midrule
    what is a walk for someone who likes climbing uphill&\texttt{grade}, \texttt{total\_gain}\\
    \midrule
    what is a walk with a variety of landscapes&\texttt{in\_urban}, \texttt{in\_woodland}, \texttt{in\_national\_parks}, \texttt{along\_surfacewater}, \texttt{along\_coast}\\
    \midrule
    what is a walk for someone seeking greater challenges&\texttt{length\_m}, \texttt{grade}, \texttt{total\_gain}\\
    \midrule
    what is a walk for someone who enjoys town walks&\texttt{in\_urban}\\
    \midrule
    what is a walk for someone who is interested in nature&\texttt{in\_urban}, \texttt{in\_woodland}, \texttt{in\_national\_parks}, \texttt{along\_surfacewater}, \texttt{along\_coast}\\
    \midrule
    what is a walk for someone who prefers wilderness to man-made&\texttt{in\_urban}, \texttt{in\_woodland}, \texttt{in\_national\_parks}, \texttt{along\_surfacewater}, \texttt{along\_coast}\\
    \bottomrule
\end{tabular}
\end{table}

\clearpage

\section{Full results}

\includegraphics[width=17cm]{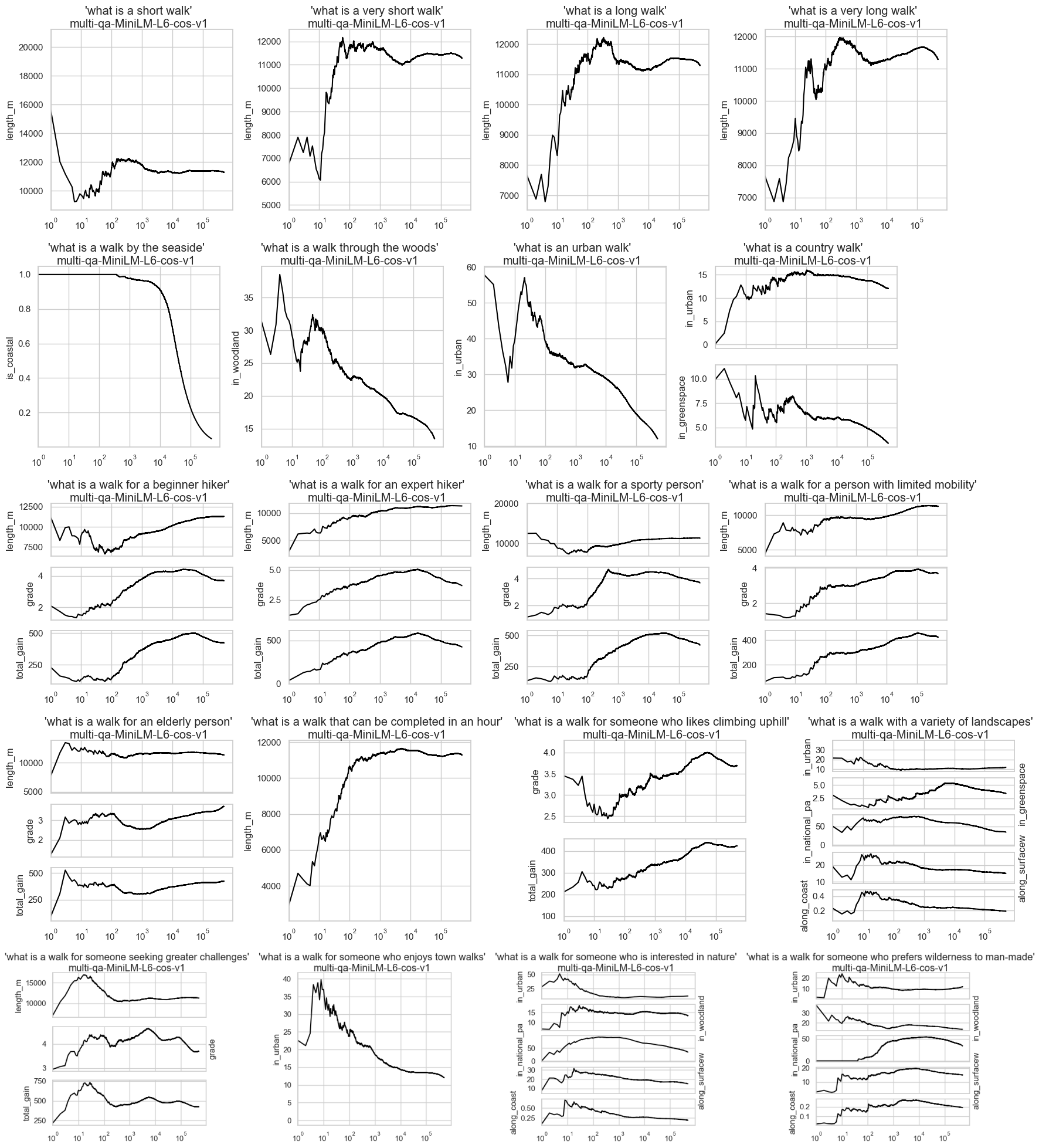}
\newpage
\includegraphics[width=17cm]{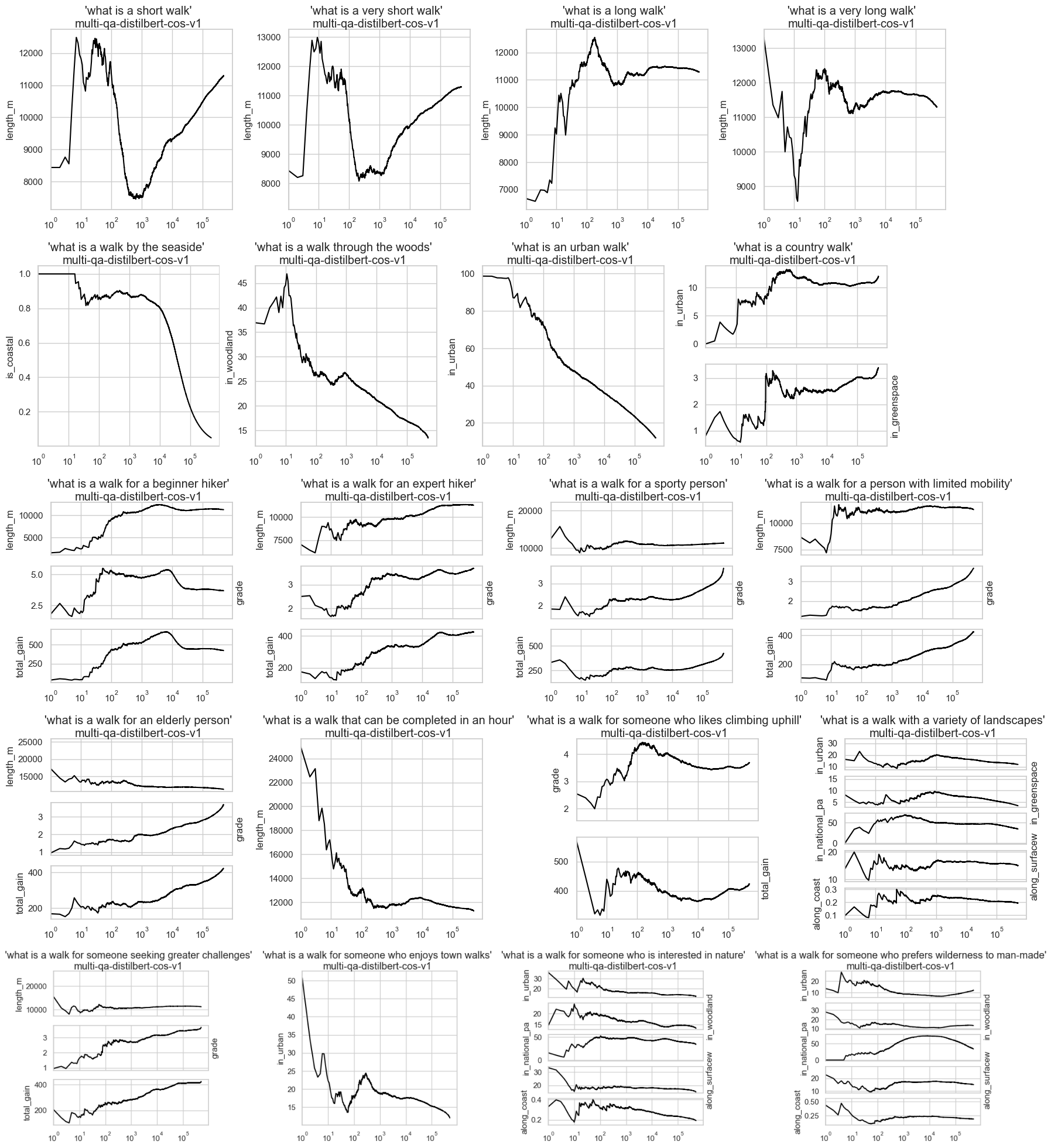}
\newpage
\includegraphics[width=17cm]{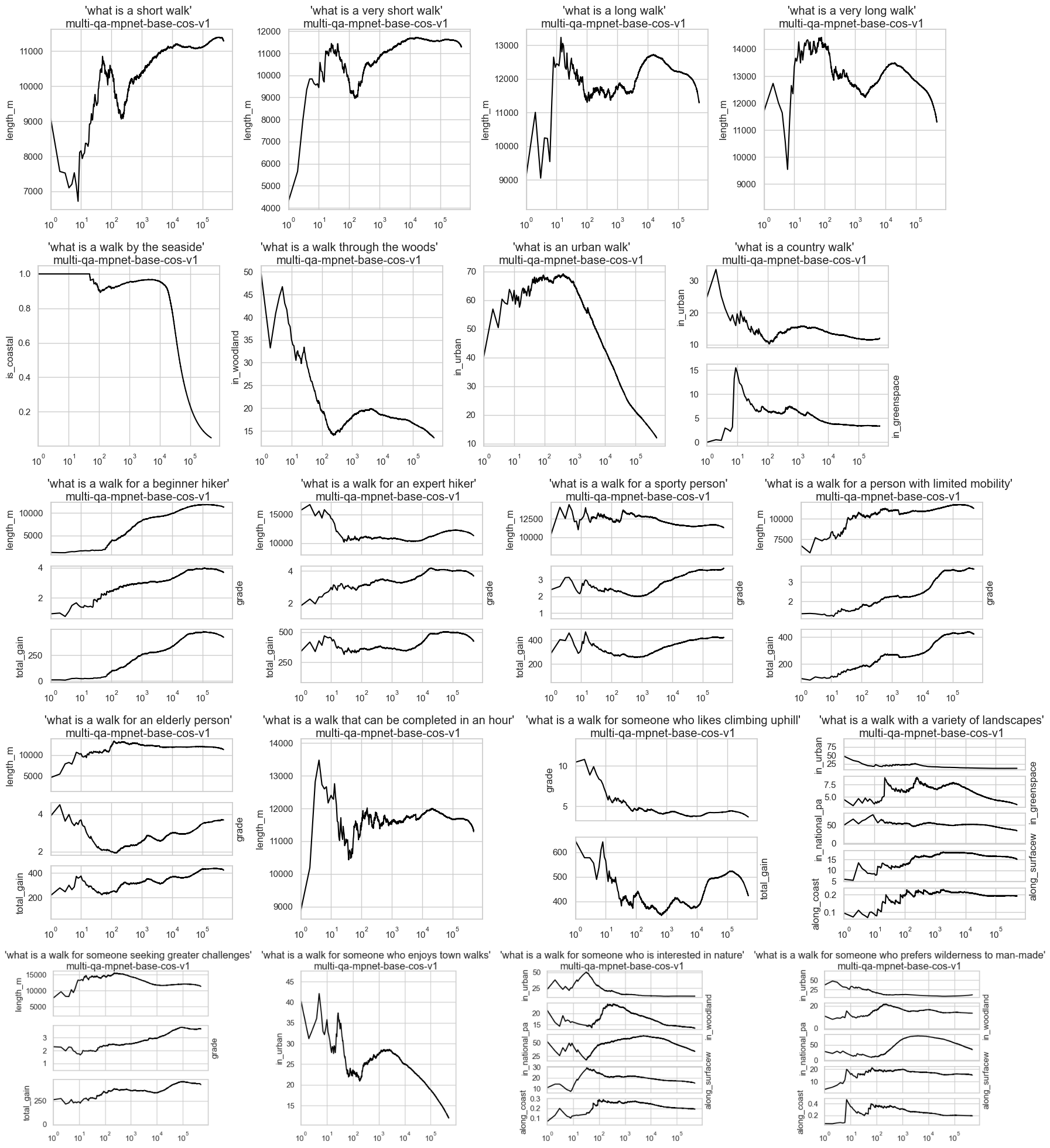}
\newpage
\includegraphics[width=17cm]{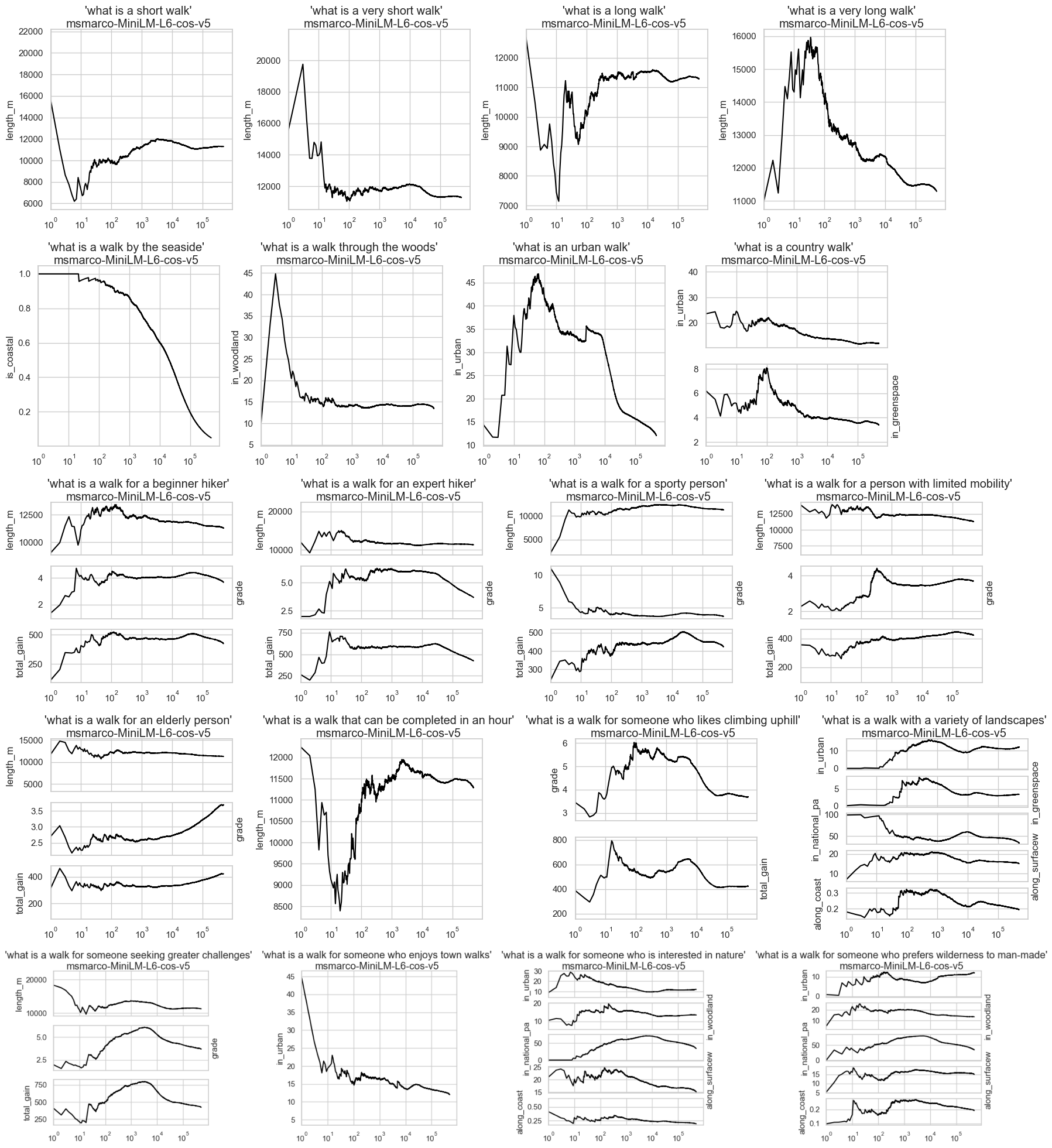}
\newpage
\includegraphics[width=17cm]{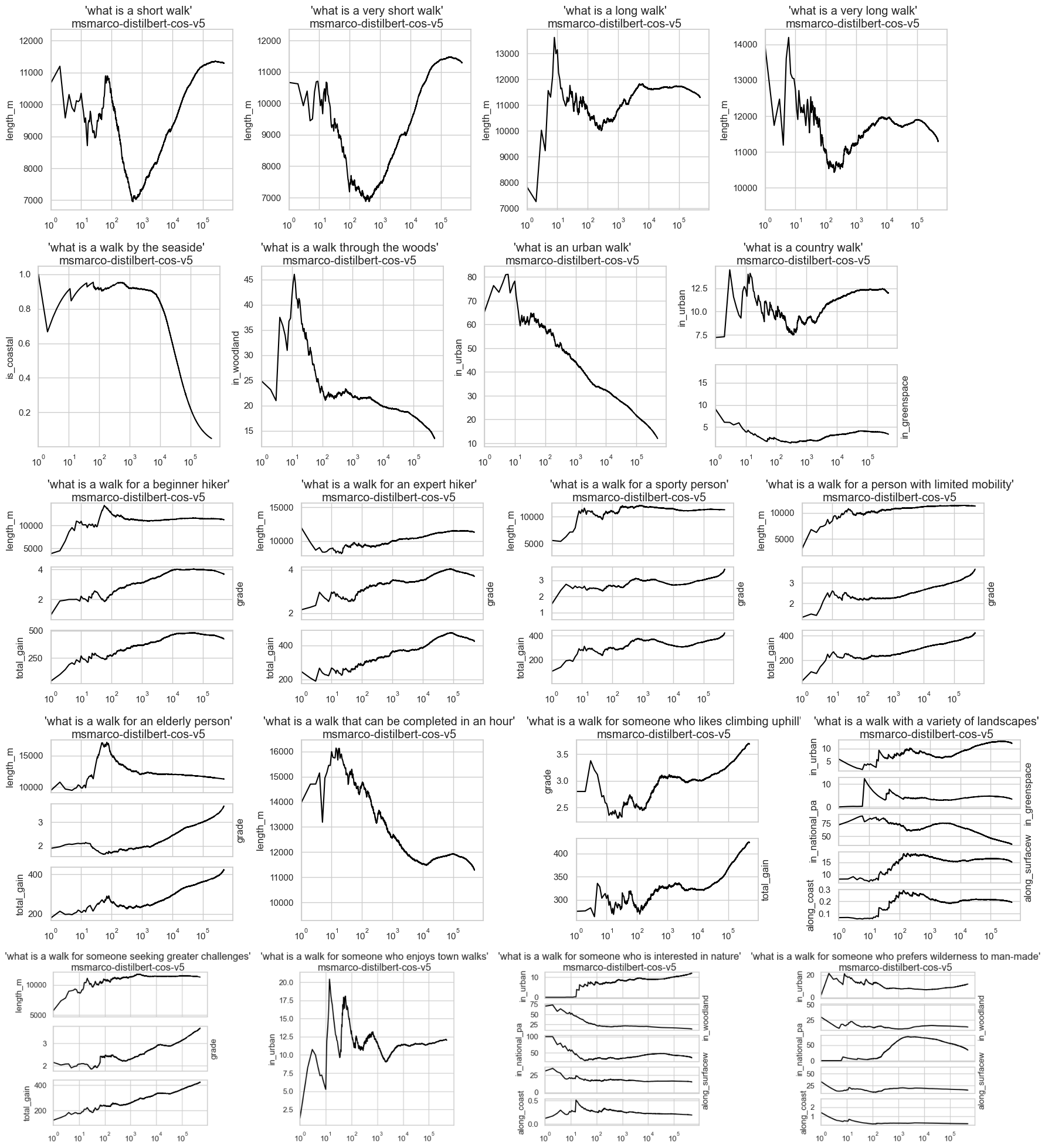}

\end{document}